\documentclass[10pt,twocolumn,letterpaper]{article}

\usepackage[pagenumbers]{cvpr} 

\usepackage[dvipsnames]{xcolor}

\definecolor{cvprblue}{rgb}{0.21,0.49,0.74}
\usepackage[pagebackref,breaklinks,colorlinks,citecolor=cvprblue]{hyperref}

\usepackage{booktabs}
\usepackage{multirow}
\usepackage{graphicx}
\usepackage{amsmath}
\newcommand{\vv}[1]{\mathbf{#1}}

\usepackage{bbding}
\usepackage{color}

\newcommand{\first}[1]{\textcolor{red}{\textbf{#1}}}
\newcommand{\second}[1]{\textcolor{blue}{\underline{#1}}}
\usepackage{bm}

\def\paperID{*****} 
\def\confName{CVPR}
\def\confYear{2024}

\begin{document}
\title{LaSe-E2V: Towards Language-guided Semantic-Aware \\ Event-to-Video Reconstruction}

\author{Kanghao~Chen$^{1}$ 
\quad Hangyu~Li$^{1}$ \quad JiaZhou~Zhou$^{1}$ \quad Zeyu~Wang$^{1,2}$
\quad Lin~Wang$^{1,2,3}$\thanks{Corresponding author}\\
$^{1}$AI Thrust, $^{2}$CMA Thrust, HKUST(GZ) \quad $^{3}$Dept. of Computer Science and Engineering, HKUST\\
{\tt\small kchen879@connect.hkust-gz.edu.cn, linwang@ust.hk} \\
\small{Project Page: \url{https://vlislab22.github.io/LaSe-E2V/}}
}



\twocolumn[{
\renewcommand\twocolumn[1][]{#1}%
\maketitle
\begin{center}
\centering
  \includegraphics[width=\textwidth]{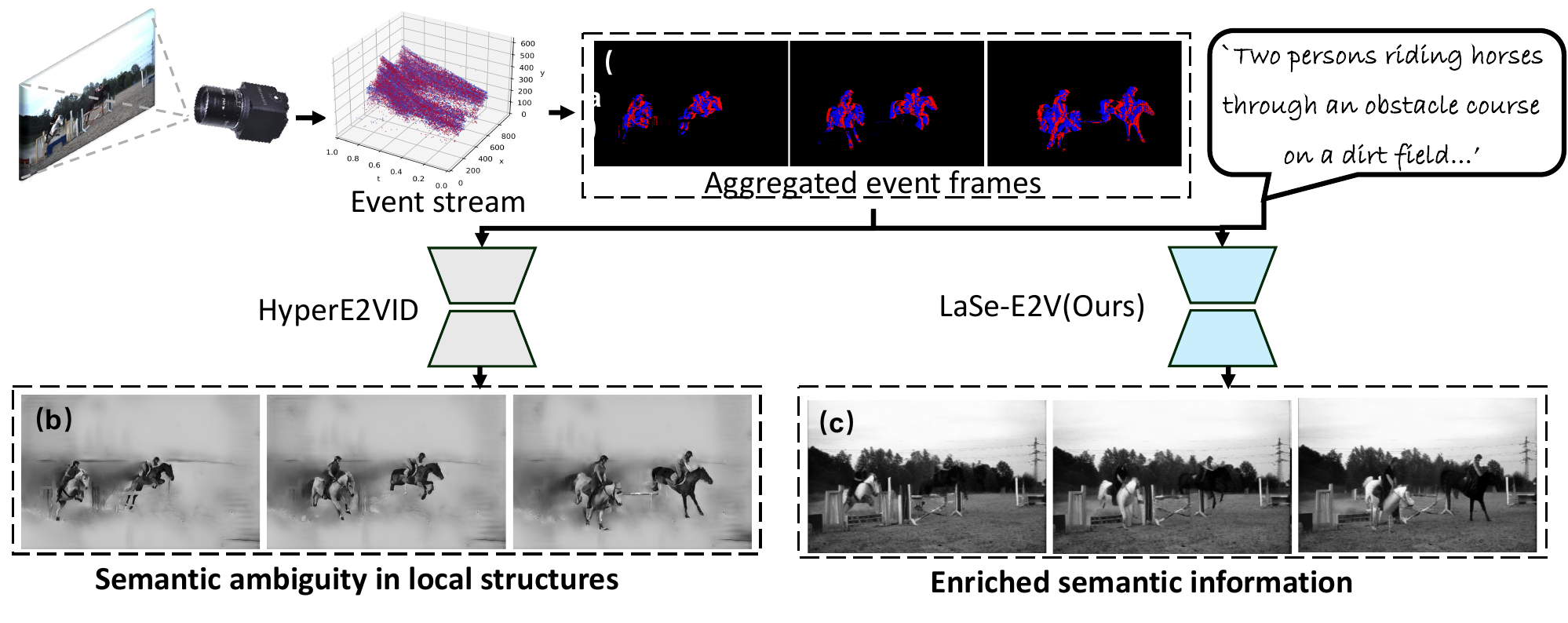}
\captionsetup{font=small}
\end{center}
\vspace{-20pt}
\captionof{figure}{
Comparison of the E2V pipeline between HyperE2VID~\cite{ercan2024hypere2vid} and our LaSe-E2V: The baseline method solely relies on event data, leading to ambiguity in local structures. In contrast, our approach integrates language descriptions to enrich the semantic information and ensure the video remains coherent with the event stream.
}
\label{fig:teaser_figure}

}]

\renewcommand{\thefootnote}{}
\footnotetext{*Corresponding author}

\begin{abstract}
Event cameras harness advantages such as low latency, high temporal resolution, and high dynamic range (HDR), compared to standard cameras. Due to the distinct imaging paradigm shift, a dominant line of research focuses on event-to-video (E2V) reconstruction to bridge event-based and standard computer vision. However, this task remains challenging due to its inherently ill-posed nature: event cameras only detect the edge and motion information locally. Consequently, the reconstructed videos are often plagued by artifacts and regional blur, primarily caused by the ambiguous semantics of event data. In this paper, \textit{we find language naturally conveys abundant semantic information, rendering it stunningly superior in ensuring semantic consistency for E2V reconstruction}. 
Accordingly, we propose a novel framework, called \textbf{LaSe-E2V},  that can achieve semantic-aware high-quality E2V reconstruction from a language-guided perspective, buttressed by the text-conditional diffusion models.
However,  due to diffusion models' inherent diversity and randomness, it is hardly possible to directly apply them to achieve spatial and temporal consistency for E2V reconstruction.
Thus, we first propose an Event-guided Spatiotemporal Attention (ESA) module to condition the event data to the denoising pipeline effectively. We then introduce an event-aware mask loss to ensure temporal coherence and a noise initialization strategy to enhance spatial consistency.
Given the absence of event-text-video paired data, we aggregate existing E2V datasets and generate textual descriptions using the tagging models for training and evaluation. Extensive experiments on three datasets covering diverse challenging scenarios (\eg, fast motion, low light) demonstrate the superiority of our method. 
\end{abstract}

\section{Introduction}
Event cameras are bio-inspired sensors that detect per-pixel intensity changes, producing asynchronous event streams~\cite{brandli2014240} with high dynamic range (HDR) and high temporal resolution. 
They particularly excel in capturing the edge information of moving objects, thus discarding the redundant visual information.
Such a distinct imaging shift poses challenges for integration with the off-the-shelf vision algorithms designed for standard cameras. To bridge the event-based and standard computer vision~\cite{jiang2019mixed,li2019event,bisulco2021fast,kendall2017geometric}, a promising way is event-to-video (E2V) reconstruction.



Recently, deep learning has been applied to E2V reconstruction~\cite{rebecq2019events,scheerlinck2020fast,weng2021event}, and remarkable performance is achieved thanks to the availability of synthetic event-video datasets and the development of model architectures.
Most existing research emphasizes on the network design~\cite{weng2021event,ercan2024hypere2vid,scheerlinck2020fast,liu2023sensing} or high-quality data synthesis~\cite{stoffregen2020reducing,liu2023sensing}.
However, this task remains challenging due to its inherently ill-posed nature: event cameras capture edge and motion information locally but neglect semantic information in regions with no intensity changes (See Fig.~\ref{fig:teaser_figure} (a)). 
Consequently, the reconstructed videos, \eg, HyperE2VID~\cite{ercan2024hypere2vid}, are plagued by artifacts and regional blur, as shown in Fig.~ \ref{fig:teaser_figure} (b).



To fill this gap, in this paper, we find language naturally conveys abundant semantic information,  which is beneficial in enhancing the semantic consistency for the reconstructed video, as shown in Fig.~\ref{fig:teaser_figure} (c). Intuitively, we propose a novel language-guided semantic-aware E2V reconstruction framework, called \textbf{LaSe-E2V}, with the text-conditional diffusion model~\cite{rombach2022high} as the backbone. 
While many efforts~\cite{wang2024videocomposer,ren2024consisti2v,ho2022video} apply diffusion models for images and video generation, adapting them to our problem is hardly possible. The reason is that the inherent randomness in diffusion sampling and the diversity objectives of generative models may result in temporal inconsistency between consecutive frames and spatial inconsistency between the event data and reconstructed videos.

Therefore, we first introduce an Event-guided Spatio-temporal Attention (ESA) module to enhance reconstructed video by introducing spatial and temporal event-driven attention layers, respectively (Sec.~\ref{sec:esa}). This module not only achieves fine-grained spatial alignment between the events and video (See Fig.~\ref{fig:ablation} (right)) but also maintains temporal smoothness and coherence by adhering to the temporal properties of event cameras. 
We then introduce an event-aware mask loss to maintain temporal coherence throughout the video by considering the spatial constraints of event data from adjacent frames (Sec.~\ref{sec:3.3Event-aware Mask Loss}). Lastly, we propose a noise initialization strategy that utilizes accumulated event frames to provide layout guidance and reduce discrepancies between the training and inference stages of the denoising process (Sec.~\ref{sec:3.4Event-aware for Noise Initialization}).

Given the absence of event-text-video paired data, we aggregate existing E2V datasets and employ tagging models to generate textual descriptions, thereby facilitating both training and evaluation processes. Extensive experiments on three widely used real-world datasets demonstrate the superiority of our method in enhancing the quality and visual effects of reconstructed videos, especially its super generalization ability when applied in challenging scenarios, e.g., fast motion, and low light (See Fig.~\ref{fig:fast_motion} and Fig.~\ref{fig:low_light}).
In summary, the contributions of our work are three-fold: (\textbf{I})  We explore E2V reconstruction from a language-guided perspective, utilizing the text-conditioned diffusion model to effectively address the semantic ambiguities inherent in event data.
(\textbf{II})  We propose the event-guided spatio-temporal attention mechanism, an event-aware mask loss, and a noise initialization strategy to ensure the semantic consistency and spatio-temporal coherency of the reconstructed video. (\textbf{III})  We have rebuilt evet-text-video paired datasets based on existing event datasets with textual descriptions generated from off-the-shelf models~\cite{zhang2023recognize}. Extensive experiments on three datasets covering diverse scenarios (\eg, fast motion, low light) demonstrate the effectiveness of our framework.

\section{Related Works}
\noindent\textbf{Event-to-Video (E2V) Reconstruction.}
E2V reconstruction falls into two categories: model-based and learning-based methods. Model-based approaches~\cite{bardow2016simultaneous,munda2018real,brandli2014real,scheerlinck2018continuous} exploit the correlation between events and intensity frames through hand-crafted regularization techniques. However, its reconstruction result is comparatively inferior to the more recent learning-based methods~\cite{rebecq2019high,rebecq2019events,scheerlinck2020fast,liu2023sensing,ercan2024hypere2vid}.
For example, E2VID~\cite{rebecq2019high,rebecq2019events} used an Unet-like network with skip connections and ConvLSTM units to reconstruct videos from long event streams. Following E2VID, SPADE-E2VID~\cite{cadena2021spade} extended E2VID to enhance temporal coherence by feeding previously reconstructed frames into a SPADE block.
HyperE2VID~\cite{ercan2024hypere2vid} introduced hypernetworks to generate per-pixel adaptive filters guided by a context fusion module. GANs, such as conditional GAN~\cite{wang2019event} and cycle-consistency GAN~\cite{yu2020event}, are used to address this issue, but often exhibit blurry images with artifacts in non-activated areas.
In summary, previous learning-based methods can merely achieve plausible reconstructed results for this problem because event data only captures motion information without semantic context information. In this work, \textit{we explore the possibility of incorporating textual descriptions with semantic awareness for E2V reconstruction.}



\noindent\textbf{Text-to-Video Diffusion Models.}
The success of Diffusion models~\cite{nichol2021glide,rombach2022high} in generating high-quality images from text prompts has advanced text-to-image (T2I) synthesis.
Inspired by T2I synthesis, text-to-video (T2V) diffusion models, such as ~\cite{blattmann2023align,harvey2022flexible,ho2022imagen,ho2022video,khachatryan2023text2video,luo2023videofusion,singer2022make,wang2023modelscope,yang2023diffusion} adapt T2I synthesis to video.
Subsequent developments~\cite{ho2022video, blattmann2023align,guo2023animatediff} focused on refining the temporal information interaction, such as optimizing temporal convolution or self-attention modules for motion learning.
Meanwhile, one T2V research line focuses on enhancing controllability by incorporating additional conditions, thus addressing the text prompts' ambiguity in motion, content, and spatial structure.
For high-level video motion control, studies propose learning LoRA~\cite{hu2021lora} layers for specific motion patterns~\cite{guo2023animatediff,zhao2023motiondirector}, or utilizing extracted trajectories~\cite{yin2023dragnuwa}, motion vectors~\cite{wang2024videocomposer}, pose sequences~\cite{ma2024follow} to guide the synthesis. 

For fine-grained spatial structure control, methods like Gen-1~\cite{esser2023structure}, VideoComposer~\cite{wang2024videocomposer}, and others~\cite{mou2024t2i,zhang2023adding} leverage monocular depth, sketch, and image control models~\cite{mou2024t2i,zhang2023adding} for flexible and controllable video generation~\cite{chen2023control,guo2023animatediff,khachatryan2023text2video,zhang2023controlvideo}.
Formally, given a video sample $x_0$, the latent diffusion model (LDM)~\cite{rombach2022high} first encodes it into a latent feature $z_0 = \vv{E}_I(x_0)$. A noisy input is obtained based on the forward diffusion process by introducing Gaussian noise to the latent representation: $x_t = \sqrt{\bar{\alpha}_t} x_0 + \sqrt{1 - \bar{\alpha}_t} \epsilon$, where $t=1,...,T$ and $T$ denotes the maximum timestep. $\bar{\alpha}_t = \prod_{i=1}^t (1 - \beta_i)$ is the coefficient that controls the noise strength. The diffusion model is trained by predicting the noise $\epsilon$ through a mean squared error: 
\begin{equation}
\label{epsilon_loss}
    l_{\epsilon} = \| \epsilon - \epsilon_{\theta}(x_t, \vv{c}, t) \|^2,
\end{equation}
where $\theta$ denotes the parameters of the U-Net of the diffusion model. $\vv{c}$ denotes the addition conditions (text, image, depth, \etal) to control the diffusion process.

These methods achieve fine-grained controllability, focusing on flexible user instructions and diverse outcomes. In this study, \textit{we employ a basic T2V diffusion model with the 3D-UNet~\cite{wang2023modelscope} architecture as the foundation for language-guided E2V reconstruction}. \textit{Our framework prioritizes video fidelity by aligning motion details from event data and semantic insights from text, thus outperforming previous models, especially in extreme scenarios like fast motion and low light.}
\section{The Proposed LaSe-E2V Framework}
\label{sec:method}

\begin{figure*}[t!]
    \centering
    \includegraphics[width=1\linewidth]{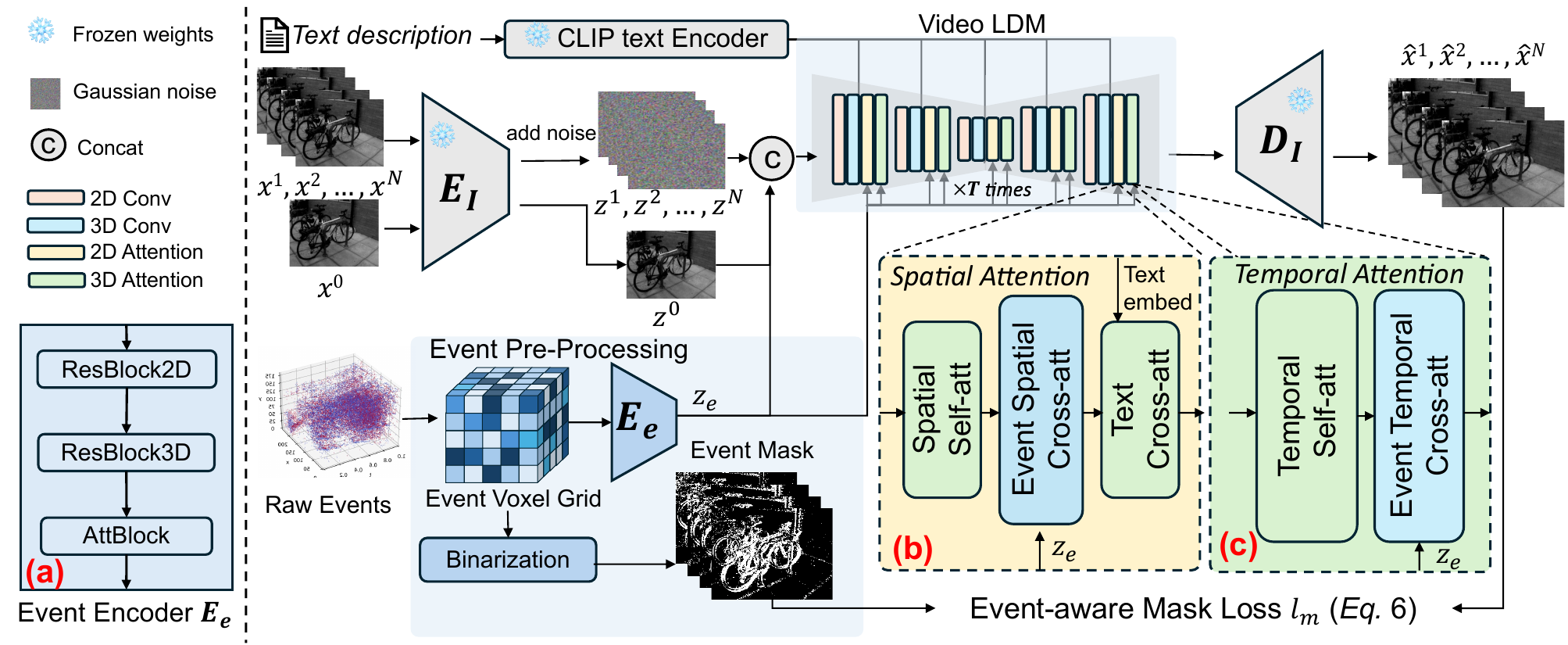}
    \caption{An overview of our proposed LaSe-E2V framework. 
    }
    \label{fig:framework}
\end{figure*}


\noindent \textbf{Event Representation.}
An event stream $\mathcal{E} = \{e_{t_k} \}^{N_e}_{k=1}$ consists of $N_e$ events. Each event $e_{t_k} \in \mathcal{E}$ is represented as a tuple $(x_k, y_k, t_k, p_k)$, which denotes the pixel coordinates, timestamp, and polarity. To make the event stream compatible with the pre-trained diffusion model, we convert $\mathcal{E}$ into a grid-like event voxel grid $V \in \mathbb{R}^{B\times H \times W}$ with $B$ time bins using temporal bilinear interpolation~\cite{zhu2019unsupervised}.

\subsection{Overall Pipeline} 

The goal of our LaSe-E2V is to reconstruct a video {$\hat{\vv{x}}=\{\hat{x}^1,\hat{x}^2,...,\hat{x}^N\}\in \mathbb{R}^{N\times C\times H\times W}$} from the set of event segments $\{\mathcal{E}^i\}^N_{i=1}$, where $\mathcal{E}^i$ is $i$-th event segment. The reconstructed video is expected to retain the motion and edge content from the event data and compensate for the semantic information from the language description.
The difficulty of our task lies not only in achieving high visual quality that aligns with the text descriptions but also in maintaining content consistency with the event data. We address the difficulty by integrating the diffusion model with events and text descriptions effectively.
As shown in Fig.~\ref{fig:framework}, LaSe-E2V consists of four major components: an image encoder $\vv{E}_I$, an event encoder $\vv{E}_e$, a video LDM, and a decoder $\vv{D}_I$.

\noindent\textbf{Image Encoder $\vv{E}_I$.}
Given a sequence of frames $\vv{x}=\{x^1,x^2,...,x^N\}$, we extract the latent representation $\vv{z}=\{z^1,z^2,...,z^N\}$ with the pre-trained image encoder of LDM~\cite{rombach2022high}.

\noindent \textbf{Event Encoder $\vv{E}_e$.}
To adapt the event voxel grids to the latent space of $\vv{E}_I$, we first project it into a latent representation with an event encoder $\vv{E}_e(\vv{V})$. 
As shown in Fig.~\ref{fig:framework}(a), we initially apply lightweight spatial 2D and temporal 3D convolution blocks to the input voxels $\vv{V}=\{V^i\}^N_{i=1}$. These blocks are designed to extract local spatial and temporal feature information, which is then processed through an attention block (\textit{AttBlock}) for global temporal modeling. Subsequently, we concatenate the event latent representation with the noise latent representation along the channel dimension to serve as a rough condition for the model. Specially, we construct the input latent representation as $\bm{\hat{\epsilon}}=\{[\epsilon^i,z_e^i]\} \in \mathbb{R}^{N \times (C'+C_e) \times H' \times W'}$, where $\epsilon^i$ and $z_e^i$ denote the noise latent and event feature corresponding to the $i$ frame with channel dimensions $C'$ and $C_e$, respectively.

\noindent \textbf{Video LDM.} We first extend the capabilities of the text-conditional image-based LDM by incorporating temporal layers in the U-Net, following the approach of video diffusion models~\cite{esser2023structure,ho2022video,singer2022make,wang2024videocomposer}.
As depicted in Fig.~\ref{fig:framework}, our framework presents a video LDM with multiple blocks consisting of 2D spatial convolution, 3D temporal convolution, 2D spatial attention, and 3D temporal attention layers. A pre-trained CLIP~\cite{radford2021learning} text encoder is employed to extract text descriptions conditioning on the attention layers to provide semantic information for the reconstructed video.
To facilitate fine-grained control of event data, we propose an Event-guided Spatio-temporal Attention (ESA) module (see Fig.~\ref{fig:framework}(b) and (c)) following $\vv{E}_e$ to convert the event voxel grids to the latent space. The ESA module enhances the control by introducing a specific spatial and temporal cross-attention in the U-Net. The technical details will be discussed in Sec.~\ref{sec:esa}. 

\noindent \textbf{Decoder $\vv{D}_I$.}
Finally, we apply the pre-trained decoder $\vv{D}_I$ of \cite{rombach2022high} to convert the estimated latent representation $\hat{\vv{z}}$ to video $\hat{\vv{x}}$ in the image space.

\noindent \textbf{Event-aware Mask Loss} $l_m$. To optimize LaSe-E2V, we propose a novel event-aware mask loss $l_m$, in addition to the $\epsilon$-prediction loss. The details of $l_m$ will be described in Sec.~\ref{sec:3.3Event-aware Mask Loss}. 

In addition, different from previous LDM models~\cite{dai2023animateanything,ren2024consisti2v}, we introduce a noise initialization strategy to alleviate the train-test gap in the inference stage, which will be described in Sec.~\ref{sec:3.4Event-aware for Noise Initialization}.

\subsection{Event-guided Spatio-temporal Attention (ESA)}
\label{sec:esa}

Only relying on the 3D U-Net and feature concatenation is not sufficient to reconstruct a content consistency video with fine-grained control for event data, as experimentally demonstrated in Fig.~\ref{fig:ablation} (right). 
Therefore, we propose to employ the event latent representation to condition on the spatial and temporal attention to ensure fine-grained spatial alignment and temporal consistency.

\noindent \textbf{Event Spatial Attention.}
As illustrated in Fig.~\ref{fig:framework}(b), the 2D spatial attention layer in the LDM U-Net integrates a self-attention mechanism that processes each frame individually and a cross-attention mechanism connecting frames with text embedding, which follows Stable Diffusion~\cite{rombach2022high}.
Intuitively, Event data indicates abundant edge information in the spatial space, which is expected to directly constrain the structural information of the reconstructed video.
To incorporate event information in the spatial attention module, 
we concatenate the features from the event latent $z_e^i$ to the U-Net intermediate features $z$ to formulate a event-based spatial attention:
\begin{equation}
z_{out} = Softmax\left(\frac{\vv{Q} \vv{K}_{e-s}^T}{\sqrt{d}}\right) \vv{V}_{e-s},
\end{equation}
where $\vv{Q} = \vv{W}^Q {z}^i$, $\vv{K}_{e-s} = \vv{W}^K [{z}^i,{z}_e^i]_s$, $\vv{V}_{e-s} = \vv{W}^V [{z}^i,{z}_e^i]_s$. $[*]_s$ represents the concatenation operation on the spatial dimension. 
This modification ensures that each spatial position in all frames accesses comprehensive information from the corresponding event data, enabling detailed structural feature control within the spatial attention layers.

\noindent \textbf{Event Temporal Attention.} In addition to the spatial information, event data inherently represents the difference between the adjacent frames, which serves as a strong constraint. 
As shown in Fig.~\ref{fig:framework}(c), to effectively leverage the temporal constraint of the event data, we facilitate event features into temporal attention. 
Specially, given the intermediate features $z$, we first reshape the height and width dimensions into the batch dimension, forming a new hidden state {$\bar{\vv{z}} \in \mathbb{R}^{(H\times W)\times N\times C}$}. Then the event latent feature $z^i_e$ is employed to interact with the latent feature $\bar{z}$ in the temporal dimension through the temporal attention layer:
\begin{equation}
z_{out} = Softmax\left(\frac{\vv{Q} \vv{K}_{e-t}^T}{\sqrt{d}}\right) \vv{V}_{e-t},
\end{equation}
where $\vv{Q} = \vv{W}^Q {\bar{z}}^i$, $\vv{K}_{e-t} = \vv{W}^K [{\bar{z}}^i,{z}_e^i]_t$, $\vv{V}_{e-t} = \vv{W}^V [{\bar{z}}^i,{z}_e^i]_t$.
$[*]_t$ represents the concatenation operation on the temporal dimension.


\noindent \textbf{Previous Frame Conditioning.}
To ensure consistency for the whole video reconstruction, we use an autoregressive way to reconstruct the video, by conditioning the model on the last frame $\hat{x}^N$ of the previously estimated video clip  $\hat{\textbf{x}}$.
We concatenate the previous frame with the input latent representation $\hat{\bm{\epsilon}}$ on the temporal dimension as a condition to obtain $\bar{\bm{\epsilon}}=\{ [z_0^0, \hat{\epsilon}^i]\} \in \mathbb{R}^{(N+1) \times C' \times H' \times W'}$.
Due to the absence of the previous frame in the first video clip, we randomly drop the previous frame condition by applying Gaussian noise to the frame latent representation.
During inference, the first video clip is reconstructed directly from the events and text data without the previous frame. For subsequent video clips, the reconstruction process utilizes the last frame from the previous video clip as a condition to ensure the temporal coherency and smoothness.

\subsection{Event-aware Mask Loss}
 
\label{sec:3.3Event-aware Mask Loss}
Considering the noise prediction loss can not capture the event constraint on temporal coherence, we introduce an event-aware mask loss to directly supervise the inter-frame difference:
\begin{equation}
    l_m = \left\| \mathcal{M} \cdot (\hat{{z}}_0^t - \hat{{z}}_0^{t-1}) \right\|_2^2,
\end{equation}
where $\mathcal{M}=I(\vv{V})$ indicates the event motion mask obtained by setting value one for event activated area and value zero for non-activated part; $\hat{\vv{z}}_0$ represents the model’s estimated clean video latent, which can be obtained by:
\begin{equation}
    \hat{{z}}_0 = \frac{{z}_t - \sqrt{1 - \bar{\alpha}_t} \epsilon_\theta({z}_t, t, \vv{c})}{\sqrt{\bar{\alpha}_t}},
\end{equation}

Finally, we combine the noise prediction loss and the event-aware mask loss with the scaling factor $\lambda$.
\begin{equation}
    l = l_{\epsilon} + \lambda \cdot l_m.
\end{equation}

\subsection{Event-aware Noise Initialization}
\label{sec:3.4Event-aware for Noise Initialization}
During training, we construct the input latent by adding noise on the clean video latent. 
The noise schedule leaves some residual signal even at the terminal diffusion timestep $T$. 
As a result, the diffusion model has a domain gap to generalize the video during the inference time when we sample from random Gaussian noise without any real data signal.
To solve this train-test discrepancy problem, during test, we obtain the base noise by adding noise on event-accumulated frames using the forward process of DDPM~\cite{ho2020denoising}. The noise latent for frame $i$ can be expressed as:
\begin{equation}
    {z}^t_T = \sqrt{\alpha_T} {z}_{e}^t + \sqrt{1 - \alpha_T} \epsilon^i,
\end{equation}
where $\alpha_T$ denotes the diffusion factor and $\vv{z}_{e}^t=\vv{E}_I(I^{-1}+\sum_0^te^i)$. The $I^{-1}$ is the last frame of the previous estimated video clip. This approach combines the base noise with the event feature to introduce layout hints to the denoising process.

\section{Experiments}
\subsection{Datasets and Implementation Details}
\label{sec:Datasets and Implementation Details}
\noindent\textbf{Dataset.}
We train our pipeline using both synthetic and real-world datasets. For synthetic data, following prior arts~\cite{rebecq2019high,liu2023sensing}, we generate event and video sequences from the MS-COCO dataset~\cite{lin2014microsoft} using the v2e~\cite{hu2021v2e} event simulator because it ensures stable and high-quality ground truth images. To enrich semantic information, we also utilize the real-world dataset BS-ERGB~\cite{tulyakov2021time}, which includes 1k training sequences. For all datasets, we employ the off-the-shelf tagging model RAM~\cite{zhang2023recognize} to generate language descriptions.
We evaluate our model on Event Camera Dataset (ECD)~\cite{mueggler2017event}, Multi Vehicle Stereo Event Camera (MVSEC) dataset~\cite{zhu2018multivehicle} and High-Quality Frames (HQF) dataset~\cite{stoffregen2020reducing}.RAM is also used to generate tags for these datasets. \textit{See more details in Appendix.}

\noindent\textbf{Implementation Details}
Based on Stable Diffusion 2.1-base~\cite{rombach2022high},  we use a text-guided video diffusion model~\cite{ren2024consisti2v} to initialize our model, pre-trained on large-scale video datasets~\cite{bain2021frozen}.
For each training video clip, we sample 16 frames and the corresponding event streams, with an interval of $1 \le v \le 3$ frames. The input size is adapted to $256 \times 256$.
Following previous methods~\cite{weng2021event,stoffregen2020reducing}, the data augmentation strategies include Gaussian noise, random flipping, and random pause.
The model is trained using the $\epsilon$-prediction objective across all U-Net parameters, with a batch size of 3 and a learning rate of 5e-5 for 150k steps on 8 NVIDIA V800 GPUs. 

\noindent \textbf{Evaluation Metric}
The Mean Squared Error (MSE), Structural Similarity (SSIM~\cite{wang2004image}), and Perceptual Similarity (LPIPS~\cite{zhang2018unreasonable}) are used to measure image quality. Given the ambiguity of the SSIM metric definition, we reassess these methods on three datasets to standardize this metric as SSIM$^*$.
\begin{table*}[t]
    \centering
    \setlength{\tabcolsep}{4pt}
    \renewcommand\arraystretch{1.3}
    \caption{Quantitative comparison with state-of-the-art methods on both synthetic and real-world benchmarks. The best and second best results of each metric are highlighted in \first{red} and \second{blue}, respectively. To align the metric of SSIM, we evaluate the previous methods based on their pre-trained models to obtain SSIM$^*$.}
    \resizebox{1.0\linewidth}{!}{
    \begin{tabular}{c|c|cccccccc|c}
    \toprule
         Datasets & Metrics & E2VID~\cite{rebecq2019high} & FireNet~\cite{scheerlinck2020fast} & E2VID+~\cite{stoffregen2020reducing} & FireNet+~\cite{stoffregen2020reducing} & SPADE-E2VID~\cite{cadena2021spade} & SSL-E2VID~\cite{paredes2021back} & ET-Net~\cite{weng2021event} & HyperE2VID~\cite{ercan2024hypere2vid}& \textbf{LaSe-E2V (Ours)} \\
         \midrule

         \multirow{4}{*}{ECD}& MSE$\downarrow$ & 0.212 & 0.131 & 0.070 & 0.063 & 0.091 & 0.046 & 0.047 & \second{0.033} &\first{0.023} \\
    & SSIM$\uparrow$ & 0.424 & 0.502 & 0.560 & 0.555 & 0.517 & 0.364 & 0.617 & 0.655 & -\\
    & SSIM$^*\uparrow$ &0.450 & 0.459 & 0.503 & 0.452 & 0.461 & 0.415 & 0.552 & \second{0.576} &\first{0.629}
\\
    & LPIPS$\downarrow$ & 0.350 & 0.320 & 0.236 & 0.290 & 0.337 & 0.425 & 0.224 & \second{0.212} & \first{0.194}\\
    \midrule
    \multirow{4}{*}{MVSEC}& MSE$\downarrow$ & 0.337 & 0.292 & 0.132 & 0.218 & 0.138 & \second{0.062} & 0.107 & 0.076 & \first{0.055}\\
    & SSIM$\uparrow$ & 0.206 & 0.261 & 0.345 & 0.297 & 0.342 & 0.345 & 0.380 & 0.419 & -\\
    & SSIM$^*\uparrow$ &0.241 & 0.198 & 0.262 & 0.212 & 0.266 & 0.264 & 0.288 & \second{0.315} & \first{0.342}
\\
    & LPIPS$\downarrow$ & 0.705 & 0.700 & 0.514 & 0.570 & 0.589 & 0.593 & 0.489 & \second{0.476} & \first{0.461}\\
    \midrule

    \multirow{4}{*}{HQF}& MSE$\downarrow$ & 0.127 & 0.094 & 0.036 & 0.040 & 0.077 & 0.126 & \second{0.032} & \first{0.031} & 0.034\\
    & SSIM$\uparrow$ & 0.540 & 0.533 & {0.643} & 0.614 & 0.521 & 0.295 & 0.658 & 0.658&- \\
    & SSIM$^*\uparrow$ &0.462 & 0.422 & 0.536 & 0.474 & 0.405 & 0.407 & \second{0.534} & 0.531 & \first{0.548}
\\
    & LPIPS$\downarrow$ & 0.382 & 0.441 & \first{0.252} & 0.314 & 0.502 & 0.498 & 0.260 & 0.257 &\second{0.254}\\

         \bottomrule
    \end{tabular}
    }
    \label{tab:main_exp}
\end{table*}

\begin{figure*}[t!]
    \centering
    \includegraphics[width=1\linewidth]{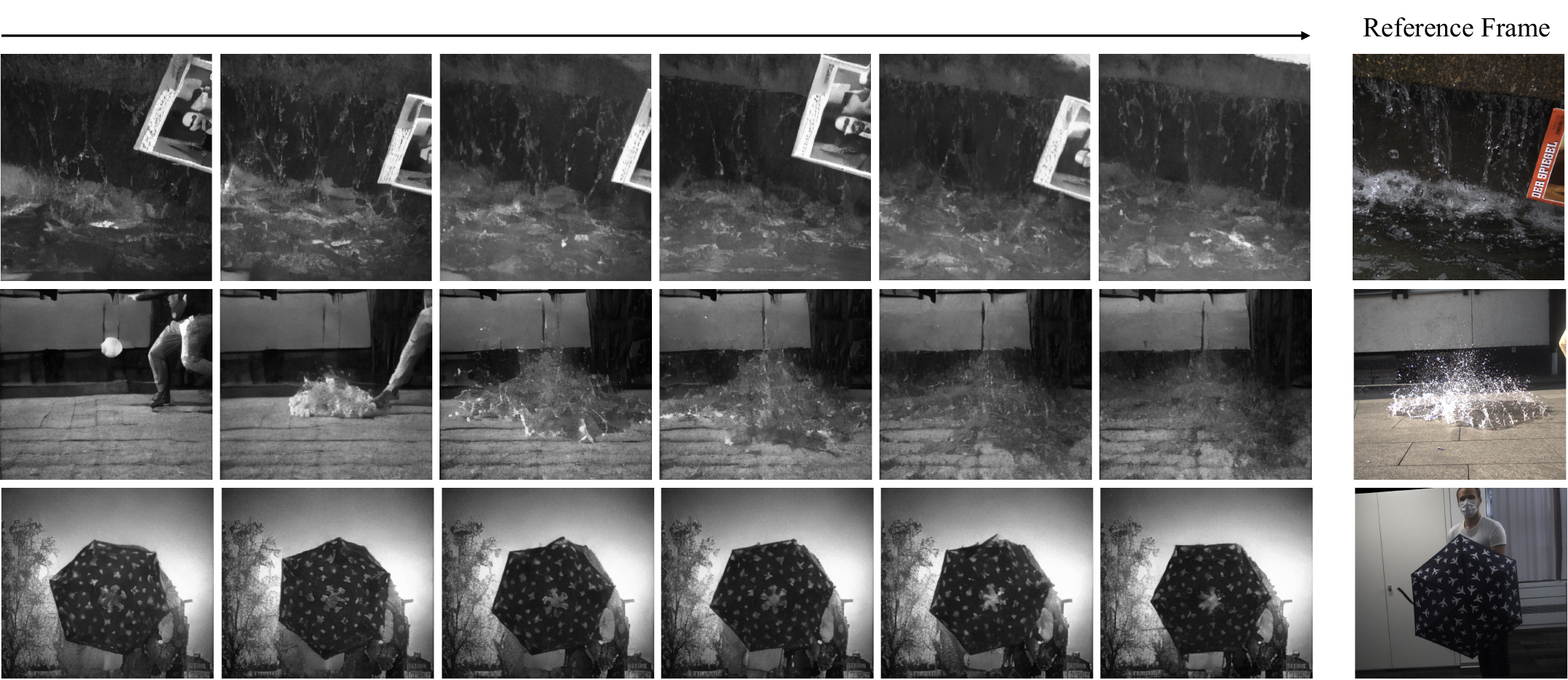}
    \caption{Qualitative results of fast-motion condition from HS-ERGB dataset~\cite{tulyakov2021time}.}
     \label{fig:fast_motion}
\end{figure*}

\subsection{Comparison with State-of-the-Art Methods}

\begin{figure*}[t!]
    \centering
    \includegraphics[width=1\linewidth]{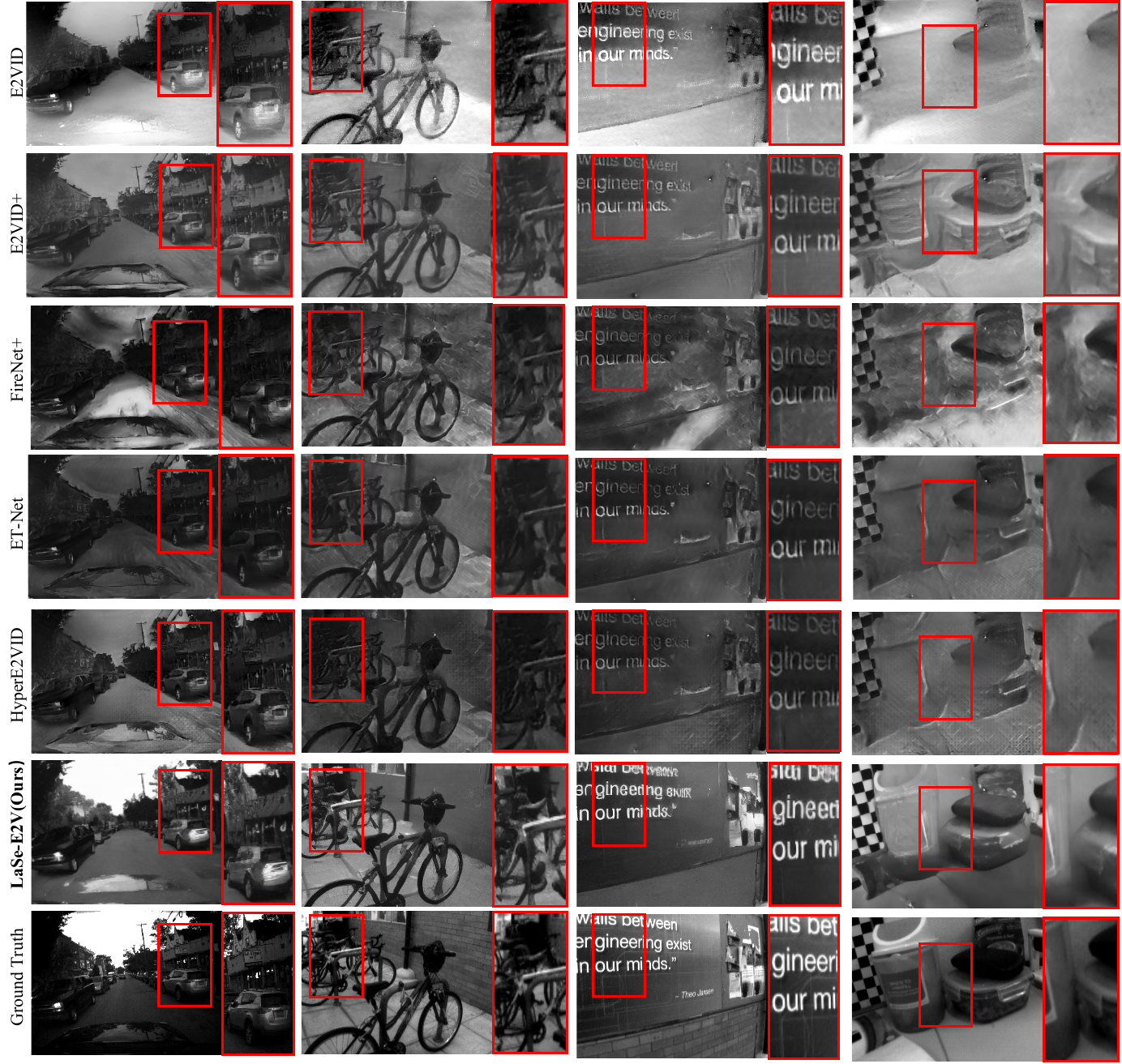}
    \caption{
    Qualitative comparisons on four sampled sequences from the test datasets. While the previous approaches suffer from low contrast, blur, and extensive artifacts, LaSe-E2V obtains clear edges with high contrast and preserves the semantic details of the objects
    }
    \label{fig:main_exp}
\end{figure*}
\textbf{Quantitative results.} We compare LaSe-E2V with eight existing learning-based methods~\cite{rebecq2019high,scheerlinck2020fast,stoffregen2020reducing, stoffregen2020reducing, cadena2021spade, paredes2021back,weng2021event,ercan2024hypere2vid}.
To ensure fair comparison, we maintain identical experiment settings without any post-processing operations across all methods.
As shown in Tab.~\ref{tab:main_exp}, LaSE-E2V mostly outperforms previous methods.
In particular, LaSE-E2V significantly surpasses HyperE2VID by $30\%$ on MSE on the ECD dataset. LaSe-E2V also excels in SSIM and LPIPS mostly. This highlights the superiority of the structural and semantic reconstruction ability of LaSE-E2V.

\noindent \textbf{Qualitative Results.} 
Fig.~\ref{fig:main_exp} illustrates the qualitative results reconstructed by our LaSe-E2V and previous methods. As shown in the street reconstruction (Column 1), our method reconstructs more details, especially for the car and the building. For the bike in Column 2, our method achieves clearer edges and higher contrast, while the previous methods are inclined to exhibit foggy artifacts around the bikes. \textbf{\textit{Please refer to the video demo and the appendix for additional qualitative results}} to demonstrate the superiority of our LaSe-E2V on temporal smoothness and consistency. 

\noindent \textbf{Results with Fast Motion. }
In Fig.~\ref{fig:fast_motion}, we show sampled reconstructed frames based on sequences of HS-ERGB~\cite{tulyakov2021time} captured by Prophesee Gen4 ($1280 \times 720$) event camera with high resolution and relatively fast motion conditions. We can see that our method is capable of clearly recovering the details for high-speed movement and preserving temporal consistency.

\begin{figure*}[t]
    \centering
    \includegraphics[width=.98\linewidth]{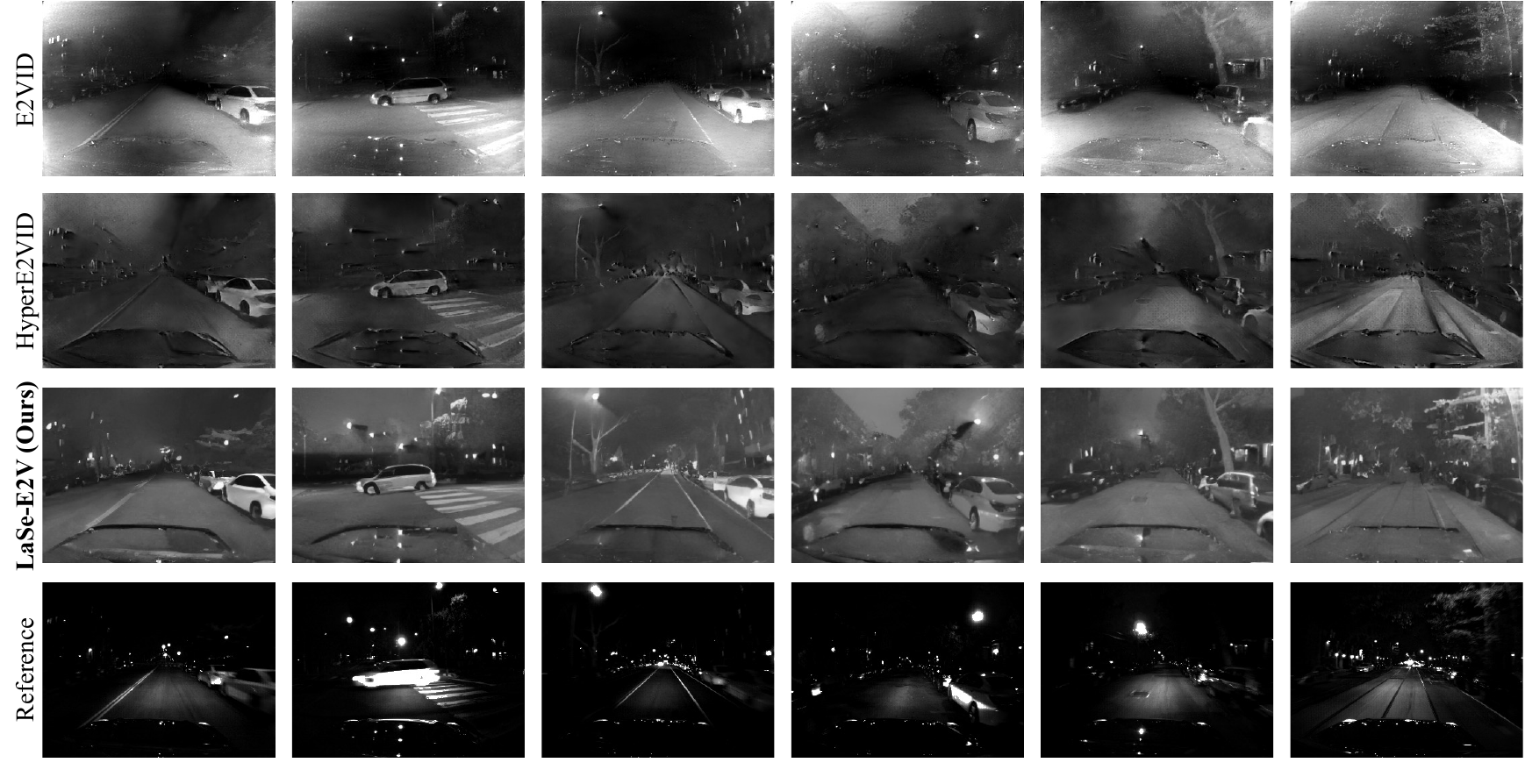}
    \caption{Qualitative results in low light condition from MVSEC dataset~\cite{zhu2018multivehicle} (\textit{outdoor\_night2}). LaSe-E2V performs better to preserve the HDR characteristic of event cameras with higher contrast.
    }
     \label{fig:low_light}
\end{figure*}

\noindent \textbf{HDR Results.}
We test our model on the video sequences in extreme conditions (\ie low light and fast motion) to further demonstrate the advantages of event cameras and the effectiveness of our framework. As shown in Fig.~\ref{fig:low_light}, we sample sequences from MVSEC (\textit{outdoor\_night2}) with relatively low-light conditions in nighty streets. 
We can see that LaSe-E2V performs better in reconstructing the scene with higher contrast and more clear edge. Compared with HyperE2VID exhibits foggy artifacts of the whole street, E2VID can reconstruct video without over-exposure or under-exposure.

\begin{figure*}[t]
    \centering
    \includegraphics[width=.96\linewidth]{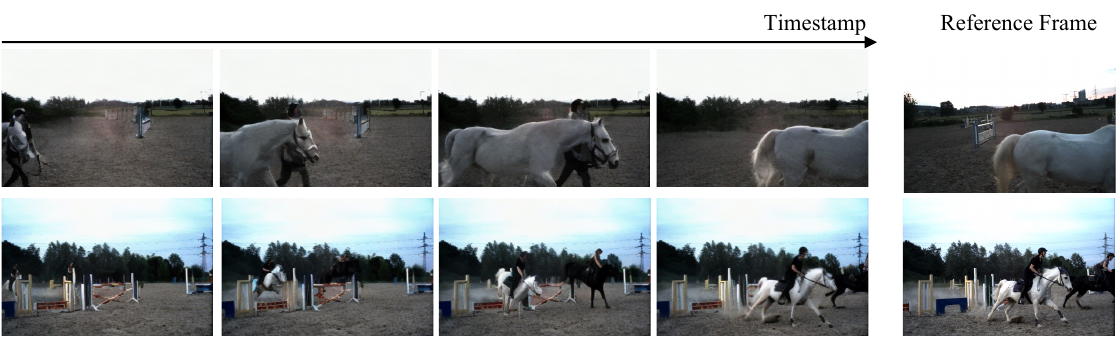}
    \caption{Qualitative results for color video reconstruction.}
    \label{fig:color_video}
\end{figure*}

\begin{figure*}[t]
    \centering\includegraphics[width=1\linewidth]{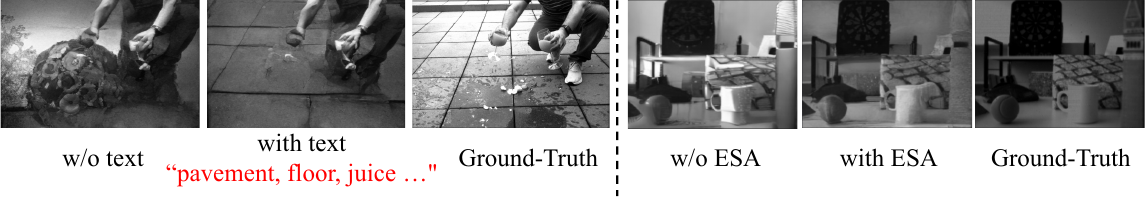}
    \caption{Qualitative comparison for the ablation study on text guidance and ESA module.}
    \label{fig:ablation}
\end{figure*}

\subsection{Discussion}

\begin{figure}
    \centering
    \includegraphics[width=1\linewidth]{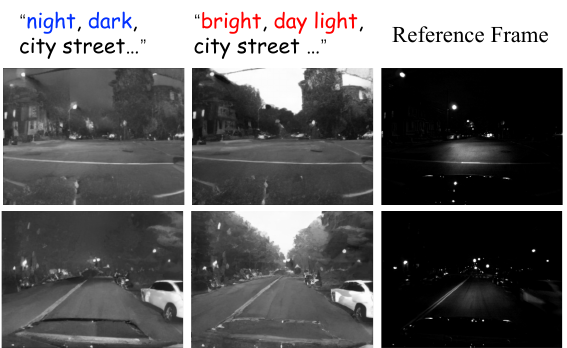}
    \caption{Qualitative results for video editing with text.}
    \label{fig:discussion}
\end{figure}

\noindent \textbf{Video Editing with Language.} In our framework, language serves as supplementary semantic information for E2V reconstruction. We investigate the impact of varying language guidance for video editing, as illustrated in Fig.~\ref{fig:discussion}. Utilizing a text description from the off-the-shelf tagging model~\cite{zhang2023recognize}, our method reconstructs a scene with a reasonable structure in low-light conditions based on descriptors like "\textit{night, dark, ...}". 


Interestingly, when we manually alter the text description to "\textit{bright, day light, ...}", the lighting condition in the scene shifts to daylight, revealing clearer details, especially in the sky area. This demonstrates our framework's ability to modify lighting conditions based on textual descriptions, by effectively modeling light conditions as semantic information.

\noindent \textbf{Color Video Reconstruction.}
Based on a pre-trained diffusion model~\cite{ren2024consisti2v}, our model inherits the capability for colored video generation by training on real-world datasets. As illustrated in Fig.~\ref{fig:color_video}, our method successfully reconstructs color videos with clear details, although it occasionally misinterprets background semantics due to the absence of event data.

\subsection{Ablation Study}
We conducted ablation experiments on LaSe-E2V framework to evaluate the effectiveness of language guidance, event-based attention, event-aware mask loss, and initialization strategy. 
\noindent \textbf{Language Guidance.} To evaluate the effectiveness of language guidance, we conducted experiments without text conditions by setting the text input to null during the denoising stage. As shown in Tab.~\ref{tab:context}, introducing the text description achieves a 0.015 improvement on SSIM.


\begin{table}
    \centering
    \caption{Ablation study on context conditions.}
    \begin{tabular}{lll|lll}
\hline
Event & Text & Frame & MSE$\downarrow$ & SSIM$\uparrow$ & LPIPS$\downarrow$ \\ \hline
\checkmark & - & \checkmark & 0.038 & 0.567 & 0.199 \\
\checkmark & \checkmark & - & 0.067 & 0.474 & 0.258 \\
\checkmark & \checkmark & \checkmark & \textbf{0.023} & \textbf{0.629} & \textbf{0.194} \\ \hline
\end{tabular}\label{tab:context}
    \label{tab:context}
\end{table}

As shown in Fig.~\ref{fig:ablation} (left), when provided with the language description "\textit{pavement}", the model tends to reconstruct the scene more closely to the ground truth, whereas the baseline model randomly generates a background which distinct to the ground truth. This demonstrates the importance of semantic information for E2V reconstruction, especially in cases where semantic ambiguity exists in the event data.
We also conduct an ablation study on the previous frame condition, as indicated in Tab.~\ref{tab:context} (row 2), where performance significantly dropped across all metrics. This underscores the critical role of the previous frame condition in our diffusion pipeline, ensuring temporal consistency within the autoregressive reconstruction process.

\noindent \textbf{Event-guided Spatio-temporal Attention.}
\begin{table}
    \centering
    \caption{Ablation study on key components."EML" denotes the event-aware mask loss. "EI" denotes event-based initialization.}
    \begin{tabular}{ccc|lll}
\hline
ESA & EML & EI & MSE$\downarrow$ & SSIM$\uparrow$ & LPIPS$\downarrow$ \\ \hline
- & \checkmark & \checkmark & 0.105 & 0.468 & 0.288 \\
\checkmark & - & \checkmark & 0.042 & 0.443 & 0.322 \\
\checkmark & \checkmark & - & 0.043 & 0.482 & 0.251 \\
\checkmark & \checkmark & \checkmark & \textbf{0.023} & \textbf{0.629} & \textbf{0.194}  \\ \hline
\end{tabular}
    \label{tab:component}
\end{table}
To demonstrate the effectiveness of our event-guided spatio-temporal attention (ESA), we conducted experiments by training the model with simple channel-wise concatenation for event input. As shown in Tab.~\ref{tab:component}, performance significantly drops without the attention mechanism, which demonstrates the effectiveness of ESA in preserving the event control on the video. Fig.~\ref{fig:ablation} illustrates that our model maintains visual content close to the ground-truth, whereas the baseline method loses control over lighting and luminance. 
We further investigate the effectiveness of the event-aware mask loss. As is shown in Tab.~\ref{tab:component} , the event-aware loss function achieves a clear margin of improvement compared to the baseline method (row 2). Moreover, Tab.~\ref{tab:component} (row 3) also demonstrates the effectiveness of our event-based initialization.

\section{Discussion and Conclusion}

\label{sec:conclusion}

\noindent \textbf{Conclusion.} 
In this paper, we introduce LaSe-E2V, a language-guided, semantic-aware E2V reconstruction method. Leveraging language descriptions that naturally contain abundant semantic information, LaSe-E2V explores text-conditional diffusion models with our proposed attention mechanism and loss function, thus achieving high-quality, semantic-aware E2V reconstruction. Extensive experiments demonstrate the effectiveness of our innovative framework. 


\noindent \textbf{Limitations and Future Work.} 
Despite the promising performance of our method, LaSe-E2V has several limitations. First, the training datasets, comprising synthetic and limited real-world data, are inadequate for optimizing data-intensive diffusion models. Consequently, our method may reconstruct scenes with artifacts differing from the training data. Second, given that LaSe-E2V relies on the diffusion model, multiple denoising steps are required for high-quality videos, slowing the process. Future work can focus on accelerating the inference efficiency based on the recent progress of diffusion models~\cite{lu2022dpm,chen2023speed,lu2022dpm2,luo2023latent}.

\noindent \textbf{Broader Impacts.} 
Based on the event camera, our LaSe-E2V enhances the capabilities of various technologies by improving the safety and robustness of intelligent systems. 
As this technology matures, its integration into everyday devices and systems seems likely, heralding a shift in how visual data is captured and utilized across industries.

\clearpage

{
    \small
    \bibliographystyle{ieeenat_fullname}
    \bibliography{main}
}


\end{document}